# Circular tests for HSM machine tools:
# Bore machining application

**Abstract:**
 Today's High-Speed Machining (HSM) machine tool combines productivity and part quality. The difficulty inherent in HSM operations lies in understanding the impact of machine tool behaviour on machining time and part quality. Analysis of some of the relevant ISO standards (230-1998, 10791-1998) and a complementary protocol for better understanding HSM technology are presented in the first part of this paper. These ISO standards are devoted to the procedures implemented in order to study the behavior of machine tool. As these procedures do not integrate HSM technology, the need for HSM machine tool tests becomes critical to improving the trade-off between machining time and part quality. A new protocol for analysing the HSM technology impact during circular interpolation is presented in the second part of the paper. This protocol which allows evaluating kinematic machine tool behaviour during circular interpolation was designed from tests without machining. These tests are discussed and their results analysed in the paper. During the circular interpolation, axis capacities (such as acceleration or Jerk) related to certain setting parameters of the numerical control unit have a significant impact on the value of the feed rate. Consequently, a kinematic model for a circular-interpolated trajectory was developed on the basis of these parameters. Moreover, the link between part accuracy and kinematic machine tool behaviour was established. The kinematic model was ultimately validated on a bore machining simulation.

## 1. Introduction

Over the past decades, the benefits and disadvantages of high-speed machining (HSM) have been presented in detail. As a result of advances in numerically-controlled machine tools, the productivity and quality of several processes have been considerably improved, e.g. end milling at high rotation speeds has become a cost-effective manufacturing process for producing parts with high precision and a high surface quality [1], [2]. Several companies have thus decided to combine the benefits of high-speed machining with flexible machine tools.

In the automotive industry for example, the parts machined are castings, cylinder heads and gearboxes. Based on these parts, typical machining features can be extracted [3] [4] such as bores, holes, drillings, tapered holes and planes. The machining of these features involves two types of displacements: displacements along the spindle axis and plane displacements normal to the spindle axis. The work presented in this paper is focused on bore machining which involves a circular interpolation of two axes during plane displacements. High accuracy and a high surface roughness are required for bores. Nevertheless, it is difficult to determine HSM impact on the machining of the corresponding features.

In this paper, an original approach to analyse HSM machine tool behaviour during circular interpolation is presented. This approach is divided into two types of circular interpolated test types: circular displacements, and spiral displacements composed of arcs of circles. Tests results have been compared to highlight the HSM impact in terms of machine tool behaviour and part quality. Differences between programmed, simulated and measured feed rates are presented. These differences are induced by the acceleration and jerk (acceleration derivative) limitations. Moreover, some numerically-controlled unit settings can induce feed rate limitations. Arc length and the transition between two circular blocks also stand out as significant parameters in the feed rate ultimately reached.

## 2. Standard procedures

Machine tool test procedures can be classified into 7 categories [5]: environmental effects, machine performance, geometric accuracy, volumetric performance, cutting performance, machining test parts and multi-function. Some of these procedures have been developed in national or international standards. The international standards ISO 230-1998 and ISO 10791-1998 propose test procedures during circular interpolation. Two types of procedures are presented in these international standards: measurements without machining, and test parts for machining procedures.

### 2.1 Procedures without machining

Circular tests are suggested in both the ISO 230-4 [6] and ISO 10791-6 [7] standards. The procedures described in these standards are based on the circular interpolation solicitation of machine tool axes. Feed rates and diameters are also given. To analyse machine tool behaviour, performance indexes are defined and some reference data are given.

The index for circularity deviation $G$ (Fig. 1a) is defined as the minimum distance between two concentric circles around the entire trajectory. It can be evaluated as the maximum radial deviation according to the square-fit method for circles.

| Nomenclature | |
|---|---|
| *ISO standard indexes* | |
| $G$ | Index of circularity deviation |
| $F_{min}$ | Minimum radial deviation |
| $F_{max}$ | Maximum radial deviation |
| *Current kinematic parameters* | |
| $V_f$ | Instantaneous feed rate |
| $V_{fr}$ | Feed rate at a discontinuity crossing |
| $\vec{V}$ | Feed rate vector |
| $A$ | Instantaneous acceleration |
| $A_n$ | Normal acceleration |
| $A_t$ | Tangential acceleration |
| $\vec{A}$ | Acceleration vector |
| $J_t$ | Tangential jerk |
| $\vec{J}$ | Jerk vector |
| $\vec{t}$ | Trajectory tangent vector at a given point |
| $\vec{n}$ | Trajectory normal vector at a given point |
| $T$ | Time |
| *Current geometric parameters* | |
| $R_r$ | Radius of curvature of a circular block |
| $\alpha_s$ | Block start angle in the XY plane (discontinuity angular position) |
| $\alpha_e$ | Block end angle in the XY plane (discontinuity angular position) |
| $\alpha_c$ | Angular position during circular interpolation |
| *Axis capacity* | |
| $V_{mi}$ | Maximum feed rate of the $i$ axis |
| $A_{mi}$ | Maximum acceleration of $i$ axis |
| $J_{mi}$ | Maximum jerk of the $i$ axis |
| *Axis capacity feed rate limitations* | |
| $V_t$ | Attainable feed rate according to axis feed rate capacity ($V_{mi}$) |
| $V_{An}$ | Attainable feed rate according to axis acceleration capacity ($A_{mi}$) |
| $V_{Jt}$ | Attainable feed rate according to axis jerk capacity ($J_{mi}$) |
| *Numerical controller parameters* | |
| $J_{curv}$ | Curvilinear jerk set in machine tool parameters |
| $r_{jct}$ | Rate of curvilinear jerk allowed in tangential jerk |
| $r_{jcc}$ | Rate of curvilinear jerk allowed in central jerk |
| $J_{tcurv}$ | Tangential jerk allowed according to curvilinear jerk ($J_{curv}$) and $r_{jct}$ |
| $t_{cy}$ | Interpolation cycle time |
| $\delta t$ | Circular transition block crossing time |
| *Numerical controller feed rate limits* | |
| $V_{Jtcurv}$ | Attainable feed rate according to $J_{tcurv}$ |
| $V_{tcy}$ | Attainable feed rate according to $t_{cy}$ |
| $V_{st}$ | Feed rate set point processed by NCU |

The index for radial deviation (Fig. 1b) is defined as the distance between the real trajectory and the nominal trajectory. The centre of the nominal trajectory is obtained by centring measurement devices on the machine tool.

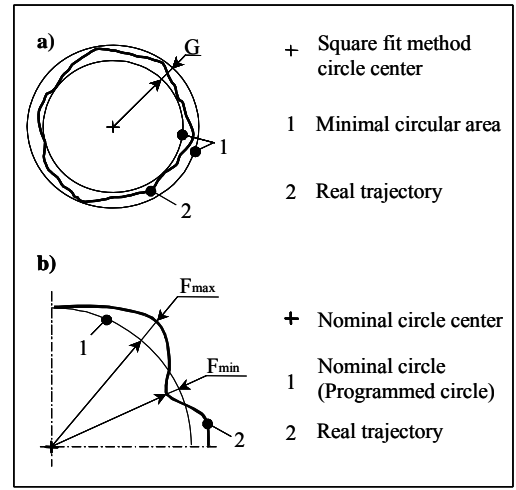

Fig. 1. a: Index for circularity deviation (G); b: Indexes for radial deviation ($F_{min}$,$F_{max}$)

The indexes and data given in ISO standards provide an efficient database for conventional machine tool behaviour. However, analysis of these indexes shows their incompatibility with HSM technology. For example, the indicated feed rates and performance indexes defined in these ISO standards correspond only to conventional machining. The proposed approach integrates new procedures compatible with HSM displacement, and the ISO 230-4 and ISO 10791-6 standard indexes.

## 2.2 Test part procedures

Test parts are advised in the ISO 10791-7 standard [8]. The first standard part is defined in order to qualify machine tool accuracy for the finishing operation of a plane. The second standard part is defined to qualify machine tool accuracy for the finishing operations of different features (Fig. 2) which are intended to control the machine tool in various displacements such as linear and circular interpolated tool paths. These test parts are specified for machine tool acceptance and may be used for comparing different machine tool structures [9].

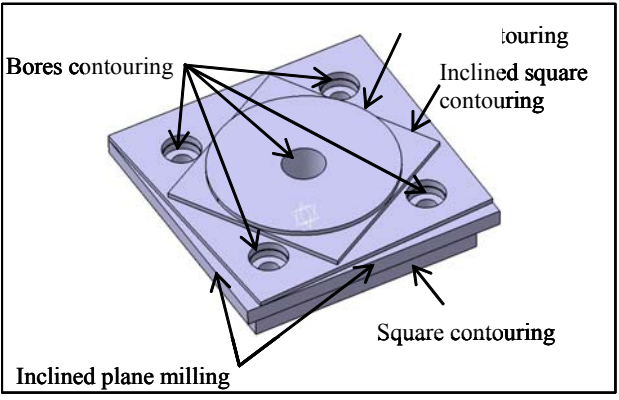

In this standard, recommendations are given for both roughing operation and cutting tool parameters which nonetheless are made available for conventional milling operations. Accuracy indexes and accuracy reference data are also provided for conventional milling.

In the standard part dedicated to circular interpolation, tests are aimed at evaluating machine tool accuracy by means of part quality. Although part quality and machining time [10] constitute the main objectives in the automotive industry, machining time has not been set in the standard.

The work presented in this paper relies upon circular interpolation associated with a test part designed for the car manufacturer RENAULT [2]. Test procedures designed in the scope of this work allow evaluating three characteristics: machine tool accuracy, machining feature quality and machining time. In addition, the link between indexes associated to these characteristics is established within an HSM context. In the following section, the specific constraints associated with HSM technology are analysed.

## 3. HSM technology

HSM technology has impacts in various machining topics, such as cutting tools [11], tool path strategies [12] [13], machine tool structures [9] [14] and numerical control units (NCU). Dugas in [15] gave a classification of the HSM technology that separates machining capacity into three fields: machine tool limits, NCU limits and cutting tool limits.

The tests presented in this paper are designed to highlight HSM performance related to the first two fields. Machine tool limits are closely associated with axes kinematics. NCU limits are introduced, in particular via the numerical axis control loop.

### 3.1 Machine tool limits

Each machine tool axis features has its own capacity. For a given axis $i$, this capacity is represented not only by the stroke, but also by kinematic parameters such as maximum feed rate ($V_{mi}$), acceleration ($A_{mi}$) and jerk ($J_{mi}$). The less dynamic axis thus imposes its kinematic skills during an interpolated displacement and then defines the limit of the feed rate. Three types of feed rate limitations are induced by axis kinematics.

According to the first limitation, the feed rate limit value ($V_t$) is determined by ($V_{mi}$), the maximum feed rate of axis $i$ that has imposed its feed rate capacity. The ($V_t$) calculation for the X- and Y-axis interpolation is detailed in Section 5.

The second limitation may be imposed by the maximum acceleration ($A_{mi}$) of the $i$ axis. At a given point on a circle trajectory with a radius $R_r$, acceleration is composed of a tangential acceleration in the tangent direction ($\vec{t}$) and a normal acceleration in the normal direction ($\vec{n}$). The unit vector of these two directions constitutes a Frenet frame ($\vec{t}$, $\vec{n}$) at the given point. Using this frame, (eq. 1) yields both the acceleration expression and the corresponding value of the feed rate ($V_f$), as defined on the tool axis.

(eq. 1) $\quad \vec{A} = \dfrac{d(V_f)}{dt} \cdot \vec{t} + \dfrac{V_f^2}{R_r} \cdot \vec{n}$ and $\quad \vec{V} = V_f \cdot \vec{t}$

When feed rate is in the steady state, the corresponding acceleration is only normal ($A_n$) and constant (eq. 1). ($A_n$) is thereby limited by the lowest maximum acceleration ($A_{mi}$) of the $i$ axis. The ($A_n$) calculation is shown in Section 5. The corresponding feed rate limitation value ($V_{An}$) is thus given by (eq. 2).

(eq. 2) $\quad V_{An} = \sqrt{R_r \times A_n}$

The third limitation case can be imposed by the maximum jerk ($J_{mi}$) of the $i$ axis. The jerk is obtained by taking the derivative of acceleration. The derivative of (eq. 1) therefore yields the expression for jerk (eq. 3). When feed rate is constant, according to (eq. 1), it can be deduced that jerk is only tangential ($J_t$), with its value given by (eq. 3). ($J_t$) is limited by the maximum jerk ($J_{mi}$) of the $i$ axis. The ($J_t$) calculation is also presented in Section 5. The corresponding feed rate value ($V_{Jt}$) is thus deduced from (eq. 3) and given by (eq. 4).

(eq. 3) $\quad \vec{J} = \dfrac{dA}{dt} \cdot \vec{n} + A \cdot \dfrac{d\vec{n}}{dt} \Rightarrow J_t = \dfrac{V_f^3}{R_r^2}$

(eq. 4) $\quad V_{Jt} = \sqrt[3]{J_t \times R_r^2}$

The feed rate ($V_s$), which equals the minimum resulting from the three limitation situations (eq. 5), is called the "static look-ahead".

(eq. 5) $\quad V_s = \min(V_t, V_{An}, V_{Jt})$

### 3.2 NCU optional functions

In the HSM context, the numerical controller unit (NCU) is intended to anticipate the trajectory in order to command axes in real time. This NCU anticipation skill depends upon its processing skill and dynamic look-ahead, which represents the number of blocks read in advance by the NCU. These two characteristics are established either by NCU construction technology or by basic machine tool settings.

The NCU tool path processing capacity is an important parameter. Trajectory anticipation and real-time command are in fact both integrated into this parameter. Moreover, this processing capacity depends on the machine tool interpolator cycle time. The trajectory is typically divided into small segments processed in accordance with this cycle time. If a segment length is too small, the NCU reduces the feed rate in order to allow enough time for axes to cover the segment with a time equal to or greater than the interpolation cycle time.

For a given arc length $R_r \times (\alpha_e - \alpha_s)$, a minimum feed rate ($V_{tcy}$) is thus computed using the interpolator cycle time ($t_{cy}$) (eq. 6).

$$\text{(eq. 6)} \qquad V_{tcy} = \frac{R_r \times (\alpha_e - \alpha_s)}{t_{cy}}$$

### 3.3 NCU specific capabilities

The realization of the tests on a high-speed machine tool imposes the use of some HSM specific modes offered by the SIEMENS 840D NCU [16] [17] used in the scope of the work. Since these HSM specific modes have an impact on the machine tool behaviour, they are defined and analysed in this section.

### 3.3.1 Continuous path mode

The tool path is locally modified at each block transition to remove tangential discontinuity. Axis displacements are smoother and machining time and cutting conditions are improved. Yet the tool path is locally modified by inserting additional curves at the block transition. For purposes of curvature discontinuity, only the acceleration jump has been authorised.

This mode is very often added with a dynamic look-ahead in order to improve efficiency. The tool path is then analysed by anticipation and an upper feed rate is generated.

### 3.3.2 Acceleration mode

The axis may be controlled using various acceleration modes: "brisk", "soft" and "drive" (Fig. 3).

In the brisk mode, axes move at maximum acceleration until the feed rate set point is reached. Machining time is optimised; however shocks in the machine tool structure are present, so part quality might be affected.

In the soft mode, axes move at constant acceleration until the feed rate set point is reached. This mode minimises shocks in the machine tool structure. Machining accuracy and part quality are better than in the brisk mode, although with increased machining time and lower productivity.

The drive mode combines both the brisk and soft modes. Axes move at maximum acceleration until reaching a feed rate limit, which has been defined within a numerical controller registry. Acceleration is then reduced according to machine tool limits until the feed rate set point has been reached.

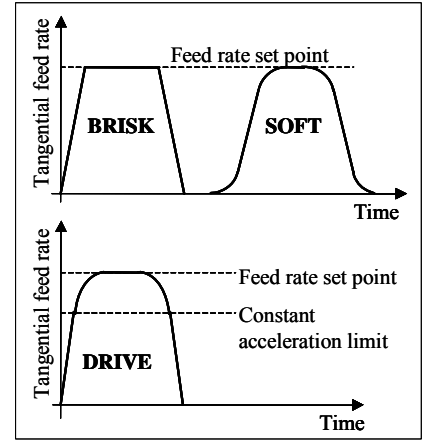

Fig. 3. Typical acceleration mode [16]

### 3.3.3 Feed forward mode

When the feed forward mode is activated, the pursuit closed-loop error is reduced to zero, in accordance with the tangential feed rate of displacement. Both displacement accuracy and part quality increase.

### 3.3.4 Experimental HSM configuration

The final HSM configuration adopted for the test integrates the continuous path mode with the tangential displacement mode (G64), the "soft" acceleration mode and the feed forward mode. This configuration has been validated from previous test results derived in [3] and those proposed by CETIM [18]. This choice is intended to achieve the fixed objectives of using machine tool structures sparingly and simulating industrial machining conditions.

## 4. Experimental framework

The protocol applied is based on tests without machining and then validated in bore machining simulations.

### 4.1 Materials

Experiments were conducted on a MIKRON UCP 710 machine tool equipped with a vertical spindle and a SIEMENS 840 D NCU. The machine tool and NCU characteristics are listed below in Table 1.

The machining simulation was performed on the test part designed for RENAULT (Fig. 4). The experiments presented in this paper have been limited to 25- and 80-mm diameter bore machining in circular interpolation.

Table 1. 5-axis HSM centre MIKRON UCP710 parameters

| Machine Tool | MIKRON UCP 710 | Spindle | Step Tec |
|---|---|---|---|
| Workspace | 710 x 550 x 500 mm | Tool holder | HSK 63A |
| Maximum feed rate ($V_m$) | 30 m/min (X,Y,Z) | Spindle Speed | 18,000 rev/min |
| Maximum | 2.5 m/s² (X) | Power | 16 kW |

| acceleration | 3 m/s² (Y) |
|---|---|
| $(A_{mi})$ | 2.1 m/s² (Z) |
| Maximum jerk | 5 m/s³ (X,Y) |
| $(J_{mi})$ | 50 m/s³ (Z) |

| Numerically-Controlled Unit | SIEMENS 840D Version 6.3 |
|---|---|
| HSM functions | Brisk, Soft, Drive, Dynamic Look-ahead, Feed forward |
| Integrated oscilloscope | Position, Feed rate, Acceleration, Spindle torque… |

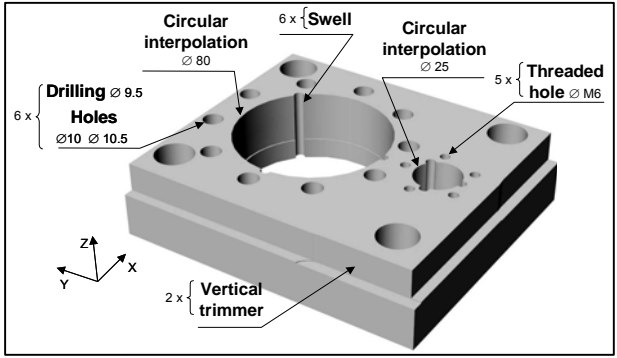

Fig. 4. Renault test part [2]

### 4.2 Performance indicators

In order to analyse tests results, three performance indicators have been defined: accuracy, kinematics and time.

For the tests carried out without machining, accuracy is evaluated from the index for circularity deviation as well as from the indexes for radial deviation defined in ISO standards (Section 2.1). This evaluation step is carried out using data processing programmes previously developed. Tool positions are measured by reading output encoder data from the machine tool-integrated oscilloscope. This type of accuracy measurement has already been validated in previous tests [3]. For part machining tests, accuracy is characterised by circular bore deviation which is evaluated using a coordinate measurement machine.

Kinematic indicators, such as accuracy counterparts, are evaluated with data stemming from the integrated oscilloscope. The associated indexes are: feed rate, acceleration and jerk.

Machining times are evaluated by means of internal clock measurements, which had previously been validated using a chronometer.

## 5. Tests without machining

### 5.1 Test protocol

In order to highlight the behaviour of machine tools without machining, three circular-interpolated trajectories following several tests were designed (Fig. 5). The first trajectory is a simple circle whereas the two others are spirals made up of half circles and quadrant. The spiral trajectories have been compared with different positions for a curvature radius variation in the X-Y plane. Various spiral inclinations have thus been tested.

Different feed rates and circle radii are tested herein to highlight the impact of radius values on machine tool behaviour (Table 2). Feed rate values have been chosen based on current industrial recommendations. For simple circular trajectories, values are chosen depending on both tool diameter (Ø 20 mm) and test part bore dimensions (Ø 25 and Ø 80 mm). The circle radii generated by tool centring come to: $R_r$=2.5 mm and $R_r$=30 mm. For spiral trajectories, a sampling interval close to the simple trajectory has been selected. This interval represents the various bore radii defined in automotive parts. Furthermore, the definition of two different curvature radius jump values allows to analyse its impact on machine tool behaviour.

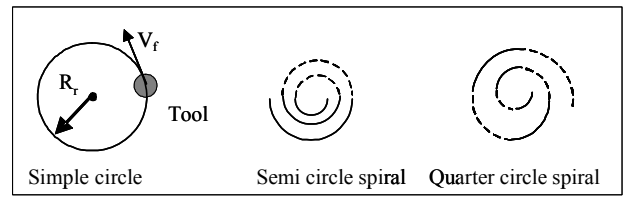

Simple circle    Semi circle spiral    Quarter circle spiral

Fig. 5. Trajectories without machining

Table 2. Test conditions

| Trajectory | Simple circle | Semicircular spiral | Quarter-circle spiral |
|---|---|---|---|
| Interpolation radius $R_r$ (mm) | 2.5 - 30 | 10 to 30 steps of 5 mm | 10 to 30 steps of 2 mm |
| Programmed feed rate (m/min) | 6 - 9 - 12 - 24 | 6 -9 - 12 - 24 | 6 -9 - 12 - 24 |

### 5.2 Simple circular trajectories

Kinematic index analyses highlight the limitation phenomena (Table 3). The programmed feed rate is not always reached and a constant feed rate is imposed. These two types of behaviour are highlighted by the measured feed rate. Circularity index analyses show the impact of machine tool kinematics, i.e. accuracy decreases as feed rate increases. In addition, the radial deviation indexes indicate the presence of a centrifugation effect, which extends over the tool axis and out of the trajectory. In order to understand these phenomena, it is proposing to simulate the static look-ahead of the machine tool and then compare the measured feed rate with the 6 m/min programmed feed rate for 2.5-mm and 30-mm circle trajectories.

Table 3. Feed rate and accuracy indexes

| Programmed feed rate (m/min) | | 6 | 9 | 12 | 24 |
|---|---|---|---|---|---|
| Measured feed rate (m/min) | ($R_r$ = 2.5) | 2.01 | 2.01 | 2.01 | 2.01 |
| | ($R_r$ = 30) | 6.01 | 9.01 | 10.53 | 10.53 |

| $R_r$ (mm) | Programmed feed rate (m/min) | G (µm) | $F_{max}$ (µm) | $F_{min}$ (µm) |
|---|---|---|---|---|
| 2.5 | 6 | 43 | 80 | -11 |
| | 9 | 43 | 80 | -11 |

| 30 | 6 | **29** | **40** | **-7.5** |
|----|----|----|----|----|
|    | 9 | **57** | **101** | **-17** |

### 5.2.1 Static look-ahead analysis

A numerical simulation of the machine tool behaviour has been carried out for both 2.5-mm and 30-mm circular interpolation in the XY plane. This simulation represents the feed rate limitation imposed by maximum axis kinematics ($V_{mi}$, $A_{mi}$, $J_{mi}$). These limitations are calculated for each angular position of the circle ($\alpha_c$).

The maximum feed rate ($V_{mx}=V_{my}$) is 30 m/min for the X and Y axes (Table 1). The feed rate limitation ($V_t$) is represented (Fig. 6) for the entire circular trajectory and then calculated for a given angular position ($\alpha_c$), using (eq. 7).

$$\text{(eq. 7)} \qquad V_t = \min\left(\frac{V_{mx}}{|\cos(\alpha_c)|} ; \frac{V_{my}}{|\sin(\alpha_c)|}\right)$$

For this complete circular interpolation ($0° < \alpha_c < 360°$), the minimum feed rate value imposed by the axes is 30 m/min in the axes direction ($90°$ and $270°$ for the X-axis, $0°$ and $180°$ for the Y-axis). In these angular positions, only one axis is interpolated; the feed rate limitation is thus limited by the maximum axis feed rate ($V_{mx}=V_{my}=30$ m/min); for other angles, $V_t$ increases. The maximum feed rate is represented in Fig. 6 and amounts to 42.4 m/min for angular positions of $45°$, $135°$, $225°$ and $315°$. For these positions, the two axes are interpolated and the limitation feed rate is given by (eq. 7):

$V_t = min[V_{mx}/\cos(45°), V_{my}/\sin(45°)] = 42.4$ m/min.

The measured feed rates (Table 3) lie beyond the scope of the simulated feed rate limits (Fig. 6). Therefore it can be concluded that axis feed rate capacities ($V_{mi}$) are not the cause of feed rate limitations measured on the machine tool evaluated. Besides, the maximum feed rate tested is less than each maximum axis feed rate, hence the limit feed rate cannot be imposed by axis feed rate capacity.

The maximum accelerations for the X and Y axes of the machine tool (Table 1) are: $A_{mx}=2.5$ m/s², $A_{my}=3$ m/s². For the entire circular trajectory at a given angular position ($\alpha_c$), the normal acceleration ($A_n$) is calculated with (eq. 8):

$$\text{(eq. 8)} \qquad A_n = \min\left(\frac{A_{mx}}{|\sin(\alpha_c)|} ; \frac{A_{my}}{|\cos(\alpha_c)|}\right)$$

The normal acceleration ($A_n$) during a circular trajectory is represented in Fig. 7 (dashed line). In this figure two minimum values for the normal acceleration can be observed. The first minimum value of 2.5 m/s² is obtained for the angular positions of $90°$ and $270°$. For these positions, only the X-axis is interpolated, thus the normal acceleration is limited by the maximum acceleration of the X-axis ($A_{mx}$). The second minimum value of 3 m/s² is reached for the angular positions of $0°$ and $180°$. For these positions, only the Y-axis is interpolated, hence the normal acceleration is limited by the maximum acceleration of the Y-axis ($A_{my}$).

The maximum normal acceleration is obtained for the angular positions of $39.8°$, $140.2°$, $219.8°$ and $320.2°$. For these angular positions, both the X- and Y-axes are interpolated with the maximum axis acceleration. For instance, the first maximum normal acceleration is obtained as follows:

$A_n = A_{mx}/\sin(39.8°) = A_{my}/\cos(39.8°)$.

This first angular position value is thus given by:

$\tan(\alpha_c) = A_{mx}/A_{my}$

The calculation can be adapted to determine each angular position for the maximum angular position.

The corresponding feed rate limited by maximum axis acceleration ($V_{An}$) is then calculated according to (eq. 2) and represented (Fig. 7, solid line) for a radius of 2.5 mm. For this simulation, it is assumed that the steady-state feed rate is reached for each angular position. The measured feed rates (Table 3) lie beyond the scope of the simulated feed rate limits (Fig. 7). It can be concluded that axis feed rate capacities ($A_{mi}$) do not cause the feed rate limitation measured on the machine tool. The same conclusion can be drawn for a 30-mm radius circle.

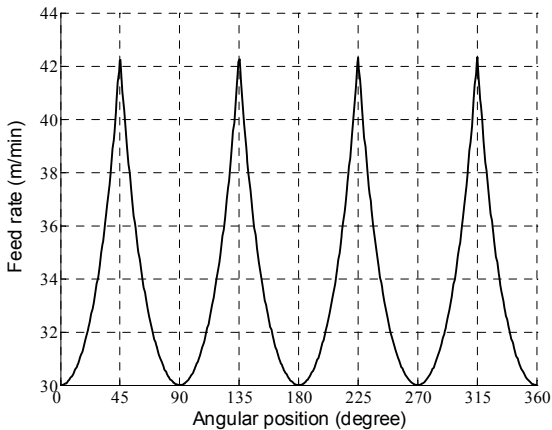

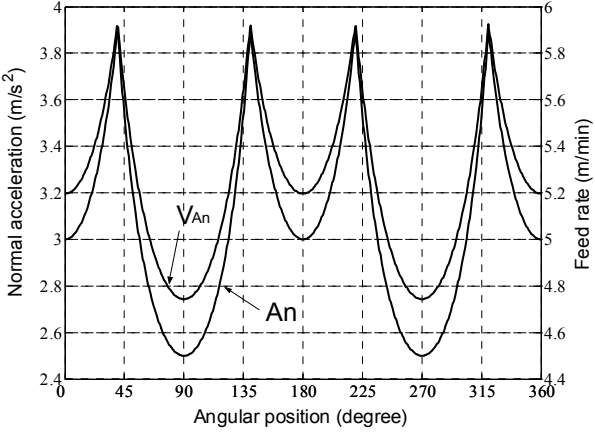

Fig. 7. Axis acceleration skills and feed rate limitation for the 2.5-mm radius

As a result of the axis jerk skills, $J_{mx}=5$ m/s³ and $J_{my}=5$ m/s³ (Table 1). The tangential jerk ($J_t$) is calculated for a given angular position ($\alpha_c$) using (eq. 9):

$$(eq. 9) \qquad J_t = \min\left(\frac{J_{mx}}{|\cos(\alpha_c)|}, \frac{J_{my}}{|\sin(\alpha_c)|}\right)$$

The tangential jerk ($J_t$) during a circular trajectory has been represented in Fig. 8 and Fig. 9 (dashed line). The minimum value of 5 m/s³ is obtained for angular positions of 0°, 90°, 180° and 270°. For the 0° and 180° positions, only the Y-axis is interpolated, hence tangential acceleration is limited by the maximum jerk of the Y-axis ($J_{my}$). For the 90° and 270° angular positions, only the X-axis is interpolated, so the tangential jerk is limited by the maximum jerk of the X-axis ($J_{mx}$).

The maximum tangential jerk is reached for angular positions of 45°, 135°, 215° and 315°. For these angular positions, both the X- and Y-axes are interpolated with maximum axis jerk. In the first angular position for example, the tangential jerk value is obtained as follows:

$J_t = J_{mx}/\cos(45°) = J_{my}/\sin(45°) = 7.07$ m/s³.

The corresponding feed rate limited by the maximum axis jerk ($V_{Jt}$) is then calculated according to (eq. 4) and represented (Fig. 8, solid line) for a 2.5-mm radius. For this simulation, it is assumed that the steady-state feed rate is achieved at each angular position. The measured feed rate limit (Table 3) of 2.01 m/min lies within the scope of the simulated feed rate limits (Fig. 8); thus axis feed rate capacities ($J_{mi}$) can cause the feed rate limitation measured on the machine tool for an interpolated circle of 2.5-mm radius. The feed rate limited by maximum axis jerk ($V_{Jt}$) is calculated from (eq. 4) for a 30-mm radius and represented in Fig. 9. The measured feed rate (Table 3) of 10.53 m/min remains within the scope of simulated feed rate limits (Fig. 9), which allows to conclude that axis feed rate capacities ($J_{mi}$) can be the cause of feed rate limitation measured on the machine tool for a 30-mm interpolated circle.

In the case of a 30-mm circle trajectory, it can be observed that programmed feed rates of 6 m/min and 9 m/min are indeed reached (Table 3). These feed rates are below the scope of the feed rate limited by maximum axis jerk: $\min(V_{Jt}) = 9.9$ m/min (Fig. 9). It is thus possible to conclude that during circular interpolation, NCU compares the programmed feed rate with the static look-ahead given in (eq. 5). When the programmed feed rate lies within the scope of a static look-ahead, NCU generates new feed rate set points.

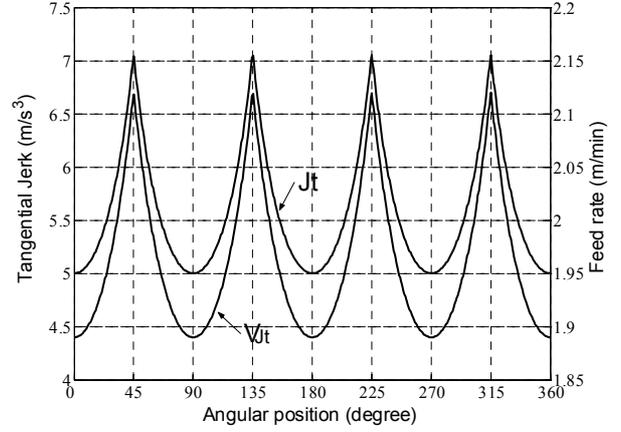

Fig. 8. Static look-ahead: Axis jerk skills for 2.5-mm radius

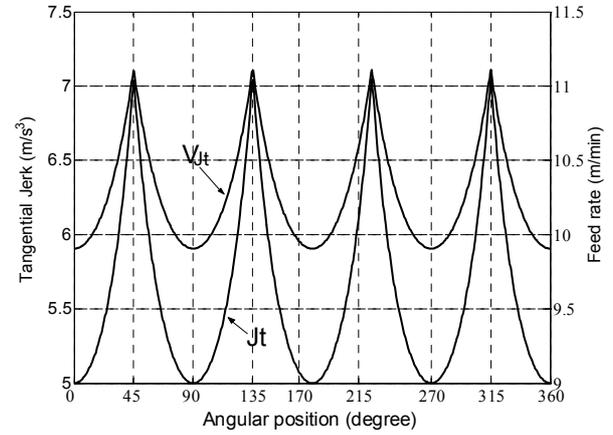

Fig. 9. Static look-ahead: Axis jerk skills for 30-mm radius

For the machine tool tested, the static look-ahead analysis shows that tangential jerk may be the cause of feed rate limitation. Each axis capacity therefore influences the generation of the feed rate law. The static look-ahead analysis however remains to be insufficient for explaining the constant imposed feed rate. To understand jerk influence and imposed constant feed rate more precisely, the feed rate law is analysed in the following section.

### 5.2.2    Kinematic law analysis

To evaluate the feed rate law influence, the records obtained from the machine tool are analysed. The feed rate and acceleration curves analysed have been recorded during one circular interpolation at 6 m/min of a 30-mm radius.

The corresponding measurements are given in Fig. 10. For this circle, the programmed feed rate has been reached. The kinematic law generated by the numerical controller is divided into three phases.

**Phase A:** The NCU generates an increase in tangential acceleration. During this phase, acceleration increases according to a linear law. The derivative of this law yields a constant jerk of 5 m/s³. The tangential acceleration increase is thus generated using the minimum tangential jerk available in a circle trajectory laid out by numerical analysis (Fig. 9): $\min(J_t) = 5$ m/s³. A tangential acceleration decrease is then generated in order to reach the programmed feed rate. This decrease is calculated by the NCU to avoid excess of speed. The tangential acceleration generated is also linear with a constant derivative of -5m/s³ equal to the minimum tangential jerk in a circular trajectory (Fig. 9): $\min(J_t) = 5$ m/s³. During phase A, the feed rate increases according to two parabolic laws (Fig. 10).

**Phase B:** The programmed feed rate is reached. The tangential acceleration and jerk become zero. Acceleration therefore is only normal (eq. 1) and constant. Using (eq. 2), this normal acceleration can be calculated $A_n = (6 \text{ m/min})^2/30 \text{ mm} = 0.333$ m/s². The survey (Fig. 10) supports the constant normal acceleration hypothesis; its value is approximately 0.335 m/s².

**Phase C:** Phase C is symmetric to phase A.

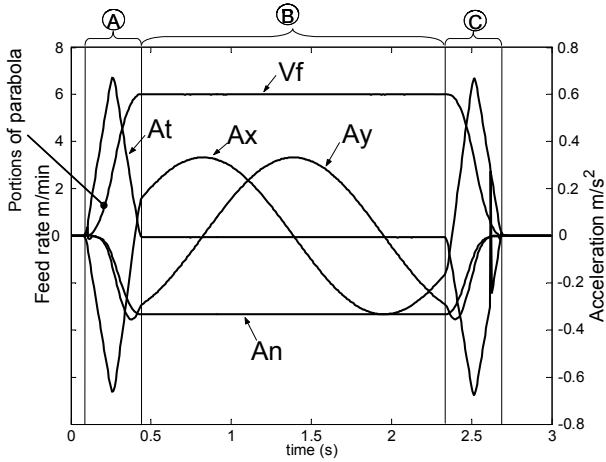

Fig. 10. Surveys conducted during 30-mm circular interpolation at 6 m/min

A second case is tested with a circle of 2.5-mm radius at 6 m/min. The corresponding measurements are shown in Fig. 11. For this circle, the programmed feed rate is not reached and remains limited to 2.01 m/min. The kinematic law generated by the numerical controller is also divided in this case into three phases.

**Phase A:** Same behaviour as in the first case.

**Phase B:** The tangential acceleration becomes zero and a constant feed rate is reached. The survey shows a feed rate of 2.01 m/min and a normal acceleration of 0.444 m/s²; these values meet (eq. 2). Therefore it can be presumed that the NCU calculates a new feed rate set point based on the static look-ahead. This new feed rate set point is lower than the simulated one represented in Fig. 7 ($\min(V_{An}) = 2.55$ m/min). Any normal acceleration limitation phenomenon occurs; this is validated by the axis acceleration presented in Fig. 11. The limit could thus be due to tangential jerk limitation, as previously predicted. During this phase, the feed rate reached is constant at 2.01 m/min (Fig. 11). This latest feed rate set point is higher than the simulated one represented in Fig. 8 ($\min(V_{Jt}) = 1.9$ m/min) and calculated to avoid the feed rate variations visible in Fig. 8. The corresponding limit tangential jerk value ($J_t$) is 6 m/s³, as given in (eq. 4). It may be now concluded that during a circle displacement, the feed rate set point is not merely being processed according to the static look-ahead ($V_s$).

**Phase C:** Same behaviour as in the first case.

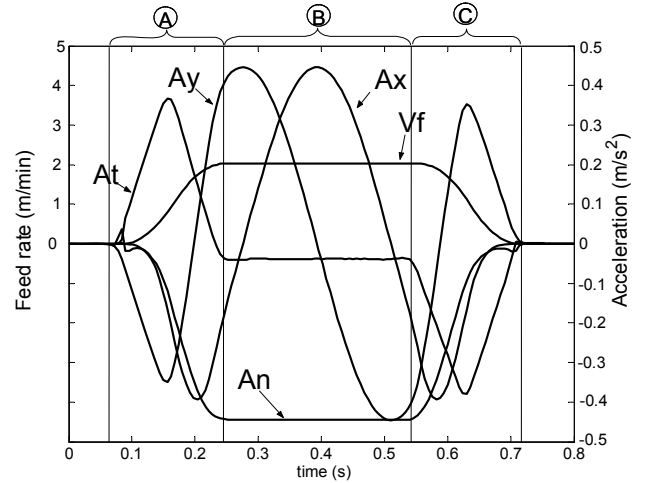

Fig. 11. Surveys conducted during 2.5-mm circular interpolation at 6 m/min

### 5.2.3 Conclusion 1: Controlled jerk behaviour

In both cases analysed, during phases A and C, the feed rate law is generated on the basis of the minimum tangential jerk reachable in the circle (Fig. 8 and Fig. 9), according to the maximum axis jerk ($J_{mi}$). During phase B, the central jerk is null, which allows to conclude that during circular interpolation, the tangential jerk value is controlled by NCU in order to generate a feed rate limited by the maximum axis jerk ($J_{mi}$).

### 5.2.4 Conclusion 2: Controlled feed rate set point

Within specific machine tool registries, a curvilinear jerk ($J_{curv} = 10$ m/s³), a rate assigned to the tangential jerk ($r_{jct} = 60\%$) and a rate assigned to the central jerk ($r_{jcc} = 40\%$) are set. These values strengthen the tangential jerk, given in (eq. 4): $J_{curv} \times r_{jct} = 0.6 \times 10 = 6$ m/s³, as determined for the 2.5-mm radius case. The feed rate is limited by the tangential jerk settings $J_{tcurv}$ (eq. 10) for the machine tool evaluated. The corresponding feed rate ($V_{Jtcurv}$) is then given by (eq. 11).

(eq. 10) $\quad J_{tcurv} = J_{curv} \times r_{jct}$

(eq. 11) $\quad V_{Jtcurv} = \sqrt[3]{J_{tcurv} \times R_r^2}$

The minimum feed rate allowed by numerical controller settings ($V_{st}$) is given by (eq. 12). ($V_{tcy}$) is the feed rate limited by interpolation cycle time, as previously presented in Section 3.2.

(eq. 12)    $V_{st} = \min\left(V_{prog}; V_{Jtcurv}; V_{Jt}; V_{An}; V_t; V_{tcy}\right)$

The feed rate set point is controlled by three main parameters: NCU settings, NC code, and axis capacities. The complete algorithm for feed rate set point processing is detailed in Fig. 12; this algorithm is based on the presence of a steady-state feed rate (phase B); i.e. axis skills are sufficient to reach the feed rate set point that has been processed according to arc length. The feed rate set point calculated in this manner must respect the axis skills represented by the static look-ahead ($V_s$) for the angular position ($\alpha_c$) when feed rate becomes constant (*Limitation on current arc of circle* in Fig. 12). As governed by NCU settings for jerk limitation, the feed rate set point reached in the steady state ($V_{st}$) is then processed (*Jerk and feed rate limitations in steady state* in Fig. 12).

For a circular trajectory of $R_r = 2.5$-mm radius at $V_{prog} = 6$ m/min: $J_{tcurv} = 6$ m/s[3] and $V_{Jtcurv} = 2.01$ m/min, it can be deduced: $\alpha_e = 0° < \alpha_c = 33.5° < \alpha_s = 360$, hence $V_{Jt}(\alpha_c) = V_{Jtcurv}$, $V_{An}(\alpha_c) = 5.7$ m/min, $V_t(\alpha_c) = 36$ m/min, $V_{tcy} = 78.5$ m/min, which yields: $V_{st} = V_{Jtcurv} = 2.01$ m/min. For a circular trajectory of $R_r = 30$ mm at $V_{prog} = 6$ m/min, according to the feed rate set point calculation flowchart: $V_{st} = V_{prog} = 6$ m/min.

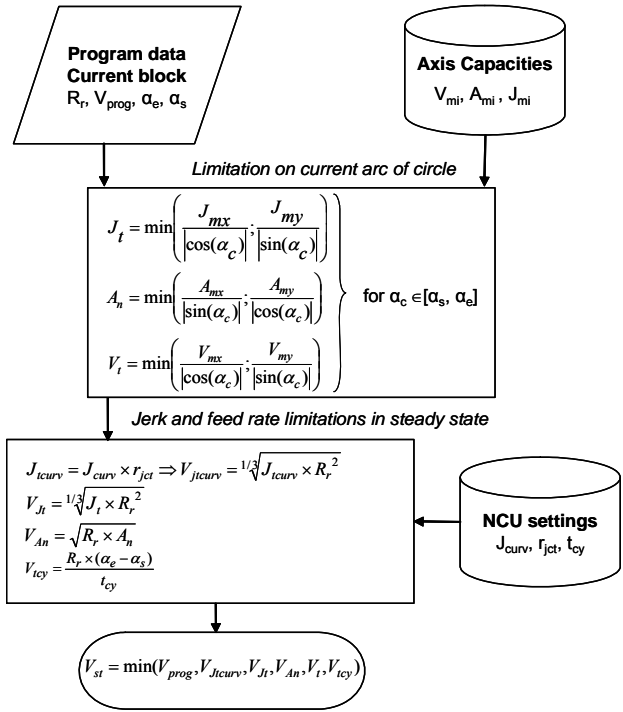

Fig. 12. Feed rate set point calculation flowchart

### 5.3    Non-inclined spiral trajectories

In this section, the circular interpolation of spiral trajectories composed of semicircles and quarter-circles are analysed. Such trajectories are performed in order to assess the influence of arc length and transition between two arcs. The semicircular spiral trajectory is defined in Fig. 13 by means of arcs (1-2), (2-3), (3-4), (4-5) and (5-6). The quarter-circle trajectory is depicted in Fig. 13 by arcs (1-1'), (1'-2), (2-2'), (2'-3), (3-3'), (3'-4), (4-4'), (4'-5), (5-5') and (5'-6).

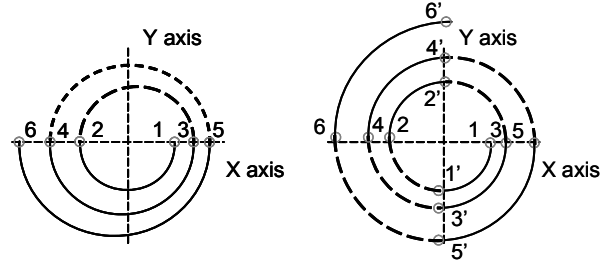

Fig. 13.  Spiral trajectories

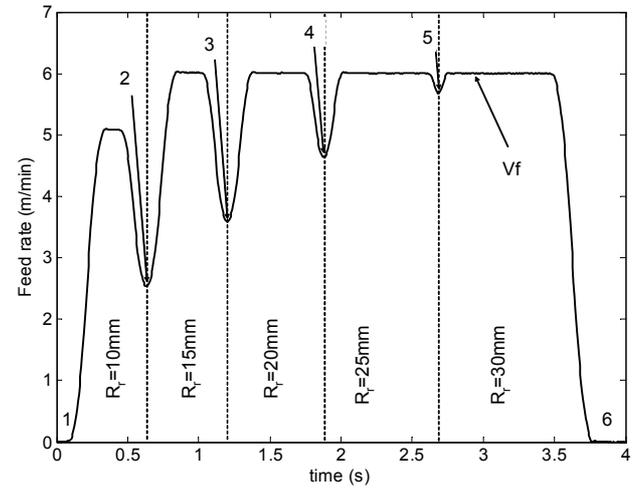

Fig. 14.  Surveys of a semicircular spiral trajectory at 6 m/min

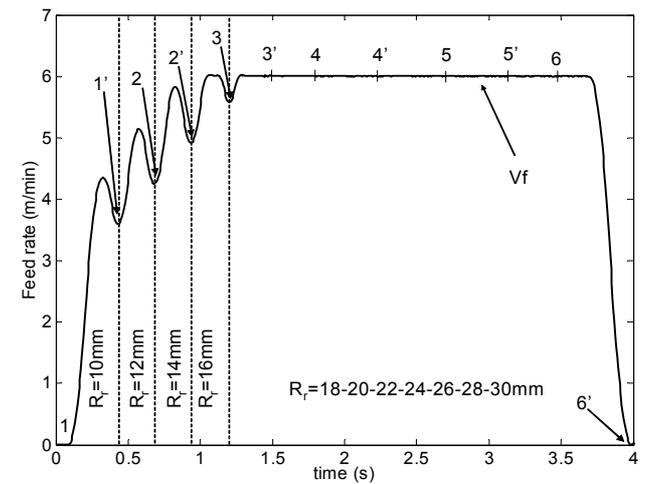

Fig. 15.  Surveys of a quarter-circle spiral trajectory at 6 m/min

### 5.3.1    Maximum arc path feed rate

When a programmed feed rate of 6 m/min is set for the semicircular spiral, for the first arc radius (10 mm),

the programmed feed rate is not reached (Fig. 14) ($V_f$ = 5.06 m/min), and a steady state with a constant feed rate is imposed. However the programmed feed rate does get reached for the three subsequent radii. These observations are confirmed by the feed rate processed using the flowchart presented in (Fig. 12).

For the quarter-circle spiral, the programmed feed rate of 6 m/min is not reached for the first arc radius (10 mm) (Fig. 15), and the constant feed rate phenomenon is not observed. There is not enough time to reach the feed rate set point. The same phenomenon is observed for 12-mm and 14-mm radii (Fig. 15). Moreover, the feed rate set point calculation (Fig. 12) yields: $V_f$ = 5.06 m/min for $Rr$ = 10 mm, $V_f$ = 5.71 m/min for $Rr$ = 12 mm, and $V_f$ = 6 m/min for $Rr$ = 14 mm. Such values are greater than the maximum feed rate reached for these radii (Fig. 15). For a 16-mm radius, the programmed feed rate is reached and a steady state (phase B) with a constant feed rate is observed. For the next radius values, the programmed feed rate is also reached and neither an acceleration phase (A) nor a deceleration phase (C) exist.

### 5.3.2 Transition block feed rate

Transitions between arcs of circle involve feed rate decreases (2,3,4,5 on Fig. 14). If the first two transitions feed rates are compared (2 and 3, Fig. 14), the feed rate is higher for the second transition. For these two transitions, the curvature radius jump remains the same (5 mm). The transition block feed rate value is thus correlated with curvature before ($R_{r1}$) and after ($R_{r2}$) the transition block.

These surveys meet the curvature discontinuity-crossing model (eq. 13) proposed in [19].

(eq. 13)   $V_{fr} = \sqrt{\dfrac{R_{r1} \times R_{r2} \times J_t \times \delta t}{|R_{r1} - R_{r2}|}}$ , where

$J_t = \min(\dfrac{J_{mx}}{\cos(\alpha)}; \dfrac{J_{my}}{\sin(\alpha)})$

The machine tool parameter δt represents the time allocated for the machine tool to cross a transition block. According to Siemens' documentation [17], the time required to cross a transition between two blocks is limited to the interpolator cycle time which is 12 ms for the machine tool tested.

Comparisons between the model (eq. 13) and the surveys result in a 0.21% mean relative error with a standard deviation of 0.22%.

In Fig. 15, after a 16-mm radius, transition blocks do not generate any feed rate loss ($V_{fr} = V_{prog}$ = 6 m/min). It can be deduced that if the block feed rate ($V_f$) before the transition is lower than the feed rate allowed for crossing the transition ($V_{fr}$), the transition does not generate any feed rate loss.

### 5.4 Inclined spiral trajectories

In the trajectories shown in Fig. 13, transitions between arcs correspond to angular positions where tangential jerk given by the maximum axis jerk ($J_{mi}$) is minimal (Fig. 9, dashed line). In order to highlight the impact of transition block placement on the feed rate generation law, each spiral trajectory analysed in the previous section is inclined. Only the quarter spiral trajectory is presented in this section as it is shown in Fig. 16 with α = 30° or 45°.

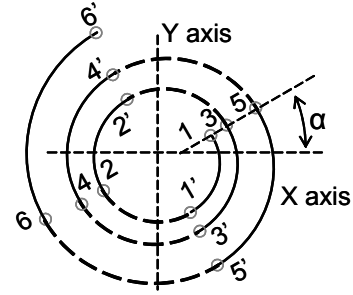

Fig. 16. Inclined spiral trajectories

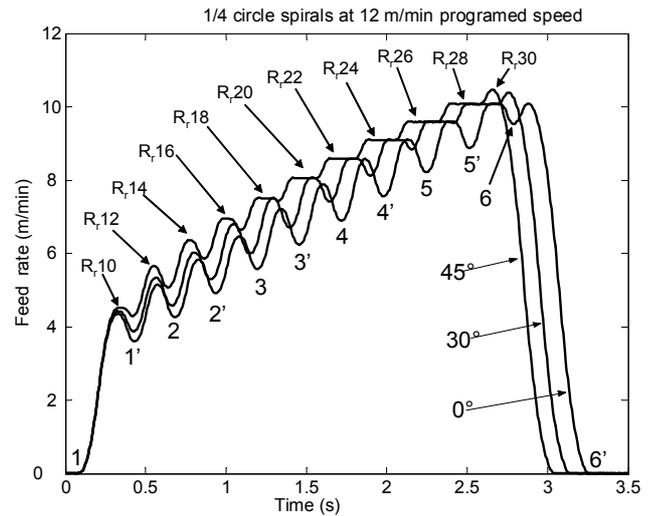

Fig. 17. Comparison of inclined spirals: quarter-circle at 12 m/min

### 5.4.1 Transition block analysis

To illustrate the transition block impact, quarter-circle spirals at 12 m/min with two different angular positions (30°, 45°) are tested and compared with non-inclined spirals (Fig. 17). For instance, the comparison between transition feed rates, for the same transition block $R_r$ = 14 mm to $R_r$ = 16 mm, yields: $V_{fr}(0°) < V_{fr}(30°) < V_{fr}(45°)$. This phenomenon can be predicted using the curvature discontinuity model defined in [19]. To cross a transition block, the feed rate is assumed to be constant, thus acceleration is normal (eq. 1). The jerk calculation (eq. 14) shows that during a transition block crossing, the jerk is tangential.

(eq. 14)   $\vec{A} = -\dfrac{V_f^2}{R_r}\vec{n}$  and  $\vec{J} = \dfrac{d(\vec{A})}{dt}$ ,

$\dfrac{d\vec{n}}{dt}$ is collinear to $\vec{t}$, hence $\vec{J} = J_t \cdot \vec{t}$; then it can be deduced (eq. 15) in the case of X and Y interpolation at a given transition block with the angular position $\alpha$.

(eq. 15)  If $\alpha \neq 0°$, then $J_t = \min\left(\dfrac{J_{mx}}{|\cos(\alpha)|}; \dfrac{J_{my}}{|\sin(\alpha)|}\right)$.

According to X- and Y-axis maximum jerk (Table 1), thus it can be written the following:
$J_t(0°) = J_{my} < J_t(30°) = 2 \times J_{my}/\sqrt{3} < J_t(45°) = J_{my} \times \sqrt{2} = J_{mx} \times \sqrt{2}$

To make the transition from a 14-mm to a 16-mm curvature radius discontinuity, a feed rate ($V_{fr}$) of 4.92 m/min is calculated (eq. 13) for 0°, a feed rate of 5.29 m/min is calculated for 30°, and a feed rate of 5.85 m/min for 45°. In the surveys conducted (Fig. 17), it is observed for 0°, 30° and 45° feed rates of 4.92 m/min, 5.29 m/min and 5.86 m/min, respectively.

### 5.4.2 Arc path feed rate analysis

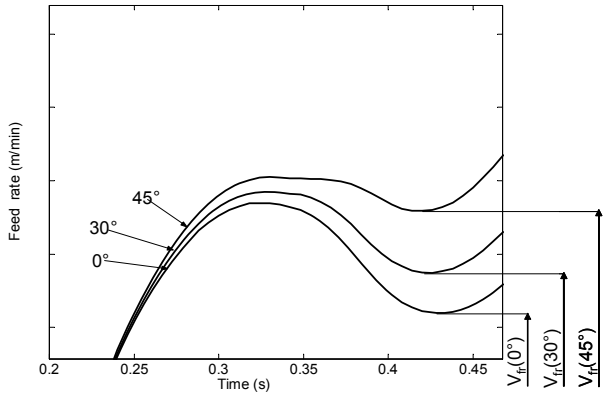

Fig. 18. 10-mm arc radius interpolation

Fig. 18 shows a close-up on the first arc radius, Rr = 10 mm in Fig. 17. For a radius of 10 mm (Fig. 17 and Fig. 18), phase B of the feed rate law has been deleted. A comparison between the maximum feed rate values reached has resulted in: $V_t(0°) < V_t(30°) < V_t(45°)$. This observation may be simply explained by the feed rate values at arc extremities. In the case analysed herein, the starting transition feed rates are the same and null, whereas the ending feed rates differ, as previously observed and explained in Section 5.4.1 by the model of (eq. 13). For a higher ending feed rate ($V_{fr}(0°) < V_{fr}(30°) < V_{fr}(45°)$), the arc length being covered to slow movement (phase C) is lower, therefore making the arc length covered to accelerate movement (phase A) greater; consequently, the feed rate reached is higher at the end of the acceleration phase (A).

### 5.4.3 Conclusion 3: Block length and block transition impact

From spiral trajectories tests impact of block length on the feed rate law have been highlighted. The presence of steady state with a constant feed rate (phase B) is detected if axis skills and arc length are sufficient to reach the feed rate set point obtained using the algorithm presented in (Fig. 12).

The curvature jump value and the angular position of the transition block both have an influence on the feed rate crossing value between two blocks. This influence is evaluated using the model described in (eq. 13). Moreover, feed rates at block start and block end are significant parameters in feed rate law generation. The presence of a steady-state feed rate (phase B) is closely correlated with these parameters and the block length. This phenomenon, which is illustrated in Fig. 17, can be observed for an arc radius of 18 mm at the block start and block end: $V_{fr}(0°) < V_{fr}(30°) < V_{fr}(45°)$. For spirals inclined at 30° and 45°, the constant steady-state feed rate (phase B) is reached, whereas for the same non-inclined spiral the steady state does not exist. In the first two cases, the arc length is sufficient to attain the feed rate set point given by the algorithm presented in (Fig. 12). However the steady state is longer for a 45° inclination than for a 30° inclination (Fig. 17). To generate the acceleration (phase A) and deceleration (phase B), the length covered is correlated with the difference between the feed rate set point ($V_f$) and the transition block feed rates ($V_{fr}(\alpha_e), V_{fr}(\alpha_s)$). This difference is higher for a 30° inclination than for 45°.

In the non-inclined spiral case, this difference is more obvious than in the other two cases. Axis skills prove insufficient to reach the feed rate set point with respect to arc length. The steady state (phase B) is therefore deleted and only the acceleration (A) and deceleration (C) phases actually exist.

### 5.5 Global kinematic behaviour

For the circular trajectory, the jerk limitation is highlighted through feed rate surveys and a static look-ahead analysis: the curvilinear jerk ($J_{curv}$) and curvilinear jerk rate, allowed up to the tangential jerk ($r_{jct}$), are used to generate the feed rate set point, as specified in the flowchart shown in Fig. 12. In addition, analysis of the feed rate law shows that the machine tool axes are controlled during circular interpolation according to the lowest of their maximum jerk ($J_{mi}$).

Complementary tests using spiral trajectories were carried out to better understand machine tool behaviour, in particular the impact of the angular position of the transition block. The influence of arc length is also highlighted in the feed rate generation law.

Finally, based on these tests, the kinematic model of the machine tool tested during circular interpolation and curvature discontinuity crossing was developed. This model implementation can be decomposed as follows:

**Stage 1: Feed rate set point calculation**

The feed rate set point is calculated from specific settings of the machine tool using the flowchart given in Fig. 12.

**Stage 2: Extremity feed rate calculation**

Circular block start and end feed rates are calculated from their angular positions in the interpolation plane. This calculation is based on the discontinuity crossing model (eq. 13) and the controlled jerk kinematic law generation given by Fig. 19.

### Stage 3: Controlled-jerk, kinematic law generation

Both the acceleration phase (A in Fig. 10 and Fig. 11) and deceleration phase (C in Fig. 10 and Fig. 11) are calculated according to arc length and extremity feed rate values (stage 1). During this stage, the existence of a constant feed rate phase (B in Fig. 10 and Fig. 11) can be deduced from the arc length. If phase B exists, the constant feed rate reached during this phase is given by stage 1, which is represented in Fig. 19. If phase B does not exist, the maximum feed rate reached ($V_f$) at the end of the acceleration phase (A) is calculated using (eq. 16). ($V_{frs}$) is the transition feed rate value between the previous and current blocks, while ($V_{fre}$) is the transition feed rate between the current and subsequent blocks.

(eq. 16)
$$\sqrt{V_f - V_{f\,re}} \times \left(V_f + V_{f\,re}\right) + \\ \sqrt{V_f - V_{f\,rs}} \times \left(V_f + V_{f\,rs}\right) = L \times \sqrt{J}$$

The acceleration phase (A) and deceleration phase (C) implemented are based on the feed-rate controlled-jerk model detailed in [20].

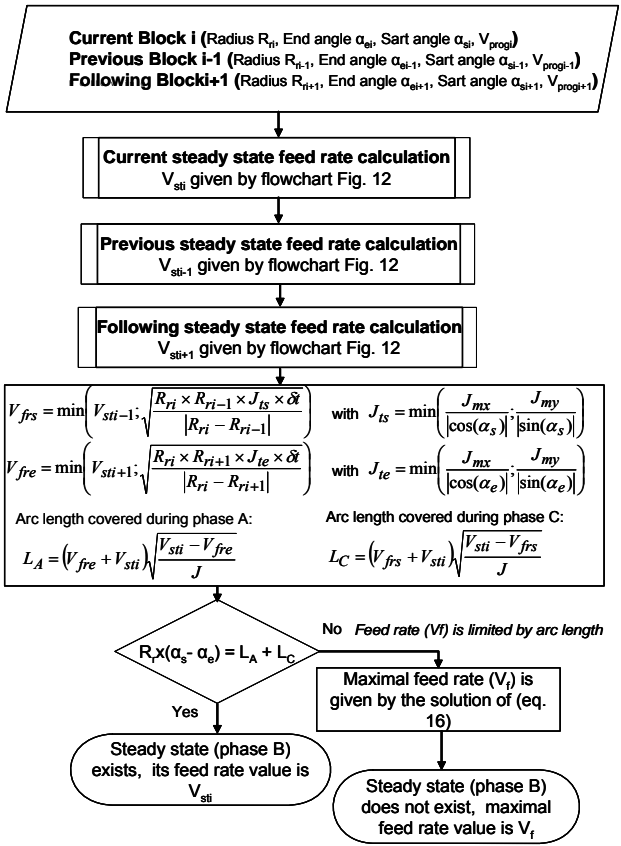

### 6. Bore-bearing machining simulation

This simulation objective is to validate the test protocol developed in the previous section and to highlight the centrifugation effect observed in Section 5.1.

The simulation has been divided into two phases: circular interpolations are first experimented without machining, then both small (Ø 25 mm) and large (Ø 80 mm) bores are machined. Simulation parameters are chosen on the basis of industrial experience (Table 4).

During these experiments, circular approaches are added to machining tool path (Fig. 20) (arcs 1-2 and 3-4). The radius value is 1.5 mm for small bores and 20 mm for large ones. Furthermore, the same circular clearances are programmed after machining.

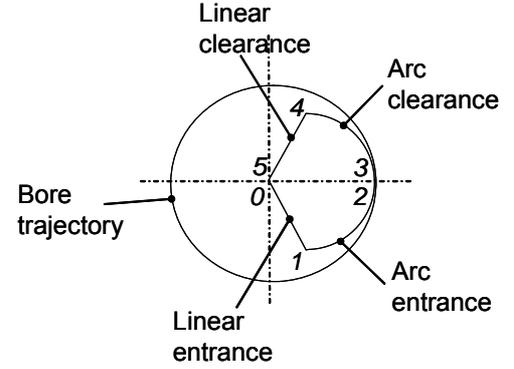

Fig. 20. Bore machining trajectory

The radial depth of cut which is 0.5 mm, is also chosen according to industrial applications. The axial depth of cut is 9 mm. The corresponding experimental results are presented in Table 5.

Table 4. Simulation parameters

| Tool | | | |
|---|---|---|---|
| Type | Carbide flat end mill | | |
| Diameter | 20 mm | | |
| Number of Tooth | 4 | | |

| Cutting conditions | | | |
|---|---|---|---|
| Features | Cutting speed | Feed per tooth | Programmed feed rate |
| Bore diameter - 80 mm | 530 m/min | 0.2 mm | 6,748 mm/min |
| | 750 m/min | | 9,549 mm/min |
| | 940 m/min | | 11,968 mm/min |
| Bore diameter - 25 mm | 470 m/min | | 5,984 mm/min |
| | 580 m/min | | 7,385 mm/min |

Fig. 19. Controlled-jerk kinematic law generation

| 670 m/min | 8,531 mm/min |

Table 5. Accuracy and time indexes

| Bore | Circularity | $T_{wm}$[1] | $T_m$[2] |
|------|-------------|-------------|----------|
| Ø 80, 530 m/min | 26 µm | 1.76 sec | 1.77 sec |
| Ø 80, 750 m/min | 56 µm | 1.36 sec | 1.35 sec |
| Ø 80, 940 m/min | 71 µm | 1.28 sec | 1.27 sec |
| Ø 25, 470 m/min | 25 µm | 0.53 sec | 0.53 sec |
| Ø 25, 580 m/min | 20 µm | 0.54 sec | 0.54 sec |
| Ø 25, 670 m/min | 20 µm | 0.53 sec | 0.53 sec |

(1): time without machining  (2): machining time

### 6.1 Small-bore machining

The feed rate path analysis shows a limitation phenomenon. For each programmed feed rate, a 2 m/min limit appears (Fig. 21). As it was previously identified in simple circular interpolation, the machine tool settings ($J_{curv}$, $r_{jct}$) impose a 2-m/min feed rate. This limited feed rate is then supported by the circularity indexes, with bore circularity being closed (Table 5). Moreover, for each test, machining time is the same: the cutting force exerts no impact on the machining feed rate.

The analysis of quarter-circle approach and clearance shows a feed rate decrease (1,2,3,4 in Fig. 21). This decrease is generated by curvature discontinuity between the approach-derived circle and the bore circle (2 and 3 in Fig. 21). The discontinuity between a 1.5 mm and 2.5 mm circle radius (approach tool path for the small bore) introduces a transition block-crossing feed rate of 0.9 m/min. This value is validated by the surveys in (1 in Fig. 21). This phenomenon is due to the tangential discontinuity between a segment tool path and the quarter-circle. A segment is in fact added between the circular approach paths and the bore centres. The phenomenon is the same for clearance tool path (4 in Fig. 21). This can be explained by activating the continuous path mode. A smoother tool path is thus generated by means of an added curve in the transition block for tangential discontinuity.

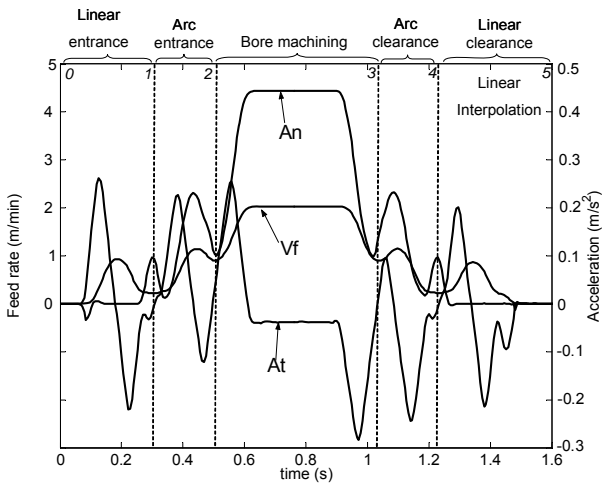

Fig. 21. 25-mm bore machining: Kinematic indexes

### 6.2 Large-bore machining

In this case, the feed rate analysis shows that the first two programmed feed rates have been reached (Fig. 22), whereas a feed rate limit of 10.55 m/min is derived for the third feed rate (Table 5). As previously determined with the simple circular interpolation, the machine tool settings ($J_{curv}$, $r_{jct}$) imposed a 10.55 m/min feed rate for a 30-mm radius circle.

The quarter-circle approach analysis highlights a feed rate decrease; this decrease is generated by a transition block between the approach-derived circle and the bore circle. The discontinuity between a 20-mm and 30-mm radius circle occurs at a feed rate crossing of 3.6 m/min, which has been confirmed by surveys. In addition, feed rate decreases are generated by the approach and the clearance tool path, as previously explained for small-bore machining.

Accuracy index analysis has confirmed the centrifugation effect. Circularity measurements show this phenomenon in bore shapes. As shown in the simple circular trajectory, a higher feed rate induces accuracy loss. Yet time indexes are better at a higher feed rate. Productivity is thus improved and the main difficulty encountered in HSM bore machining is to be able to anticipate accuracy losses as a function of cutting condition.

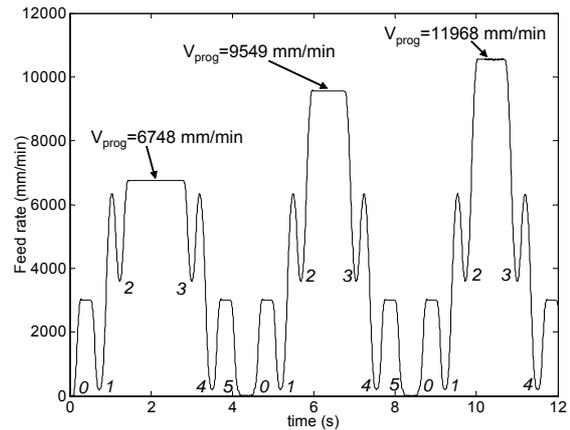

Fig. 22. 80-mm bore machining: Kinematic indexes for 3 feed rates

### 6.3 Kinematic model validation

Bore machining simulation has confirmed the phenomena highlighted in tests without machining: feed rate limitation, and centrifugation effects. Moreover, machine tool kinematic behaviour is closely correlated with part accuracy and productivity. The predicted machine tool behaviour during tool path generation thus proves to be highly significant. Machining time, feed rate loss and potential impact on part accuracy should all be analysed by means of kinematic machine tool behaviour simulation.

The model developed using tests without machining, has been validated on the basis of these tests. By means of very simple tests without machining, significant machine tool parameters can be highlighted and easily integrated into the kinematic model.

## 7. Conclusion

HSM technology influences both part quality and machining time. Experts need to strike a balance between part quality and machining time with respect to this technology. Standards are not sufficient to assist experts for solving this problem. In this context, test procedures that enhance machine tool accuracy, the machining quality and machining time evaluation must be further developed. An original approach to analyse HSM machine tool behaviour during circular interpolation is presented in this paper.

The test procedure elaborated in this approach is based on machine tool and NCU limits. In HSM context kinematic parameters (axis feed rate, acceleration and jerk) can imply feed rate limitation. Besides, NCU processing capacity and specific capabilities influence feed rate law generation and imply feed rate loss.

To emphasize the influence of these HSM parameters, test types without machining have been designed herein. Both circular and spiral trajectories have been performed and assessed. Test analyses rely upon three types of indicator: accuracy ($F_{min}$,$F_{max}$,G), kinematics (feed rate, acceleration, jerk), and time indexes.

The test procedure has been applied to the evaluation of a given machine tool. The results obtained underscore the link between part accuracy, machining time, NCU behaviour and kinematics parameters. Influence of the jerk (acceleration derivative) has been emphasised. This parameter is controlled during circular displacements by means of NC code, the numerically-controlled unit and kinematic capacities of axes. Thus, the feed rate set point processed by NCU can be lower than the programmed feed rate. Besides, a centrifugation effect has been emphasized. Accuracy indicators underline the loss of part quality when the maximal reached feed rate is higher. Thus, to maintain a given part quality, sometimes the programmed feed rate should be reduced. Besides, for a given interpolated arc radius, a circularity index limit can be induced by feed rate limitation.

Based on these tests, a kinematic machine tool model was developed. This model was then validated through bore machining and phenomena underlined during simple tests without machining were confirmed. The developed kinematic model has allowed to simulate each bore machining time and to anticipate loss of bore quality. Besides, this kinematic model has permitted to analyse respect of cutting condition.